%% file: samplepaper.tex
\documentclass[runningheads]{llncs}

\usepackage[T1]{fontenc}
\def\doi#1{\href{https://doi.org/\detokenize{#1}}{\url{https://doi.org/\detokenize{#1}}}}
\usepackage{bm}
\usepackage{amsfonts}
\usepackage[misc,geometry]{ifsym}

\usepackage{graphicx}
\usepackage{multirow}
\usepackage{subfigure}
\usepackage{booktabs}
\usepackage{color}
\usepackage{listings}
\newcommand{\TheName}{\texttt{MUSCLE}}

\usepackage[dvipsnames]{xcolor}
\usepackage{mathtools, nccmath}

\usepackage{graphicx}

\usepackage{tikz}
\colorlet{linecol}{black!75}
\usepackage{xkcdcolors} 

\usepackage{tikz}
\usetikzlibrary{tikzmark}
\usetikzlibrary{calc}
\newcommand{\highlight}[2]{\colorbox{#1!17}{$\displaystyle #2$}}

\renewcommand{\highlight}[2]{\colorbox{#1!17}{#2}}

\begin{document}
\title{MUSCLE: Multi-task Self-supervised Continual Learning to Pre-train Deep Models for X-ray Images of Multiple Body Parts}
%
%
\author{Weibin Liao\inst{1} \and
Haoyi Xiong~(\Letter)\inst{1}\orcidID{0000-0002-5451-3253} \and
Qingzhong Wang\inst{1}\and
Yan Mo\inst{1}\and
Xuhong Li\inst{1}\and
Yi Liu\inst{1}\and
Zeyu Chen \inst{1}\and
Siyu Huang\inst{2}\and
Dejing Dou\inst{1}}
%
%
%
\authorrunning{W. Liao, H. Xiong, Q. Wang, Y. Mo, X. Li, et al.}
\titlerunning{MUSCLE}
%
\institute{Baidu, Inc., Beijing, China\and
Harvard University, Cambridge, MA}
%

\maketitle              
\begin{abstract}
While self-supervised learning (SSL) algorithms have been widely used to pre-train deep models, few efforts~\cite{ref_article_1} have been done to improve representation learning of X-ray image analysis with SSL pre-trained models. In this work, we study a novel self-supervised pre-training pipeline, namely \emph{\underline{Mu}lti-task \underline{S}elf-super-vised \underline{C}ontinual \underline{Le}arning} (\TheName{}), for multiple medical imaging tasks, such as classification and segmentation, using X-ray images collected from multiple body parts, including heads, lungs, and bones. 
Specifically, \TheName{} aggregates X-rays collected from multiple body parts for MoCo-based representation learning, and adopts a well-designed continual learning (CL) procedure to further pre-train the backbone subject various X-ray analysis tasks jointly. Certain strategies for image pre-processing, learning schedules, and regularization have been used to solve \emph{data heterogeneity}, \emph{over-fitting}, and \emph{catastrophic forgetting} problems for multi-task/dataset learning in \TheName.
We evaluate \TheName{} using 9 real-world X-ray datasets with various tasks, including pneumonia classification, skeletal abnormality classification, lung segmentation, and tuberculosis~(TB) detection. 
Comparisons against other pre-trained models~\cite{ref_article_2} confirm the \emph{proof-of-concept} that self-supervised multi-task/dataset continual pre-training could boost the performance of X-ray image analysis.

\end{abstract}
\section{Introduction}
While deep learning-based solutions~\cite{ref_article_0} have achieved great success in medical image analysis, such as X-ray image classification and segmentation tasks for diseases in bones, lungs and heads, it might require an extremely large number of images with fine annotations to train the deep neural network (DNN) models and deliver decent performance~\cite{ref_article_i1} in a supervised learning manner. To lower the sample size required, the self-supervised learning (SSL) paradigm has been recently proposed to boost the performance of DNN models through learning visual features from images~\cite{ref_article_i2} without using labels.


Among a wide range of SSL methods, contrastive learning~\cite{ref_article_i3} algorithms use a similarity-based metric to measure the distance between two embeddings derived from two different views of a single image, where the views of image are generated through data augmentation, e.g., rotation, clip, and shift, and embeddings are extracted from the DNN with learnable parameters. In particular, for computer vision tasks, the contrastive loss is computed using the feature representations of the images extracted from the encoder network, resulting in the clustering of similar samples together and the dispersal of different samples. Recent methods such as SwAV~\cite{ref_article_4}, SimCLR~\cite{ref_article_5}, MoCo~\cite{ref_article_2}, and PILR~\cite{ref_article_6} have been proposed to outperform supervised learning methods on natural images. While contrastive learning methods have demonstrated promising results on natural image classification tasks, the attempts to leverage them in medical image analysis are often limited~\cite{ref_article_7,ref_article_8}.
While Sowrirajan et al.~\cite{ref_article_1}  proposed MoCo-CXR that can produce models with better representations for the detection of pathologies in chest X-rays using MoCo~\cite{ref_article_2}, the superiority of SSL for \emph{other X-ray analysis tasks, such as detection and segmentation, on various body parts, such as lung and bones, is not yet known}. 

\subsubsection{Our Contributions} 
\begin{figure}
\centering
\includegraphics[width=0.85\textwidth]{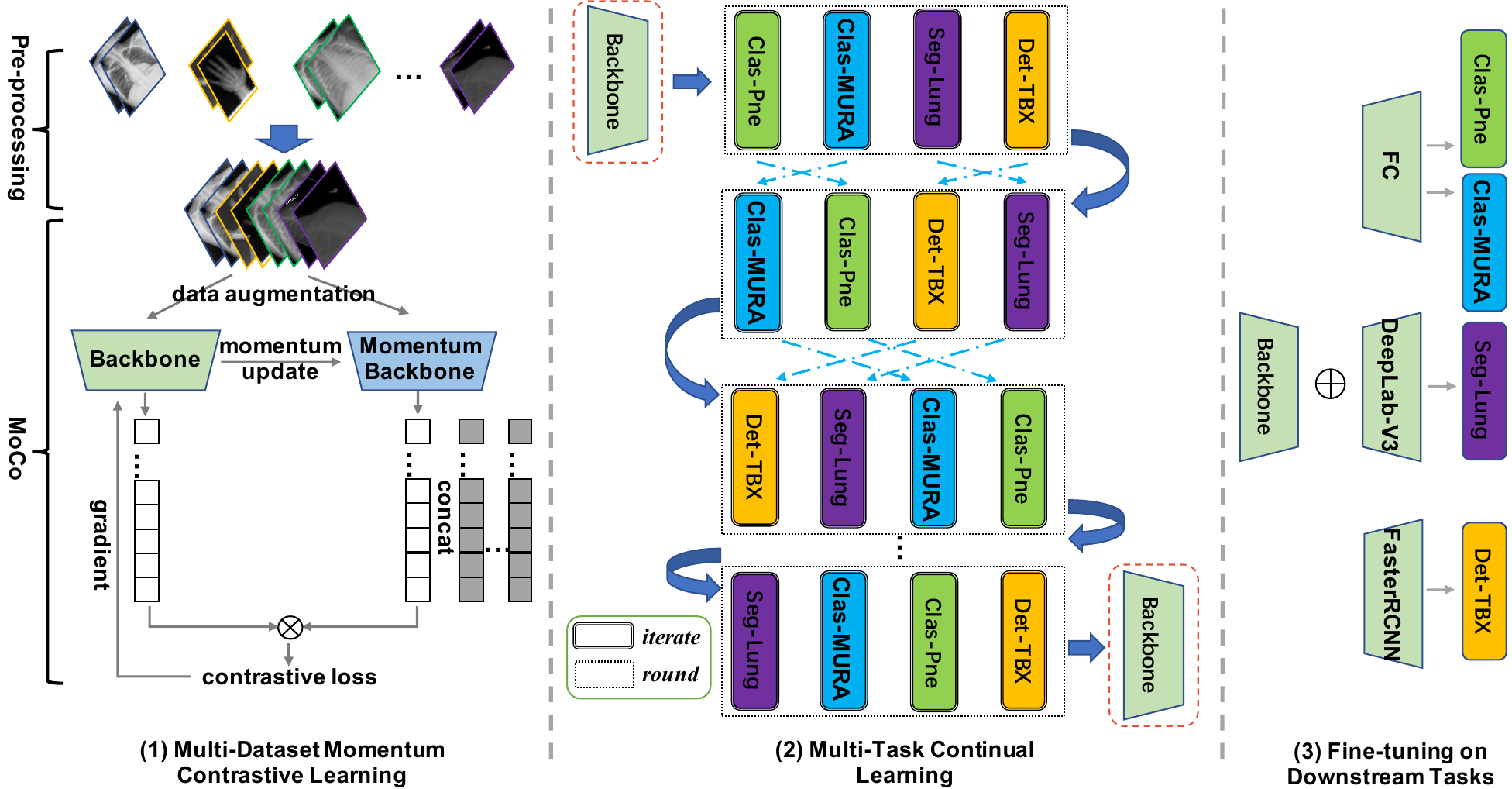}
\caption{The pipeline of \TheName{}, consists of three parts (1) Multi-Dataset Momentum Contrastive (Multi-Dataset MoCo) Learning, (2) Multi-Task Continual Learning and (3) Fine-tuning on Downstream Tasks.} 
\label{fig1}
\end{figure}

In this work, we proposed \TheName{}--\emph{\underline{MU}lti-task \underline{S}elf-supervised \underline{C}ontinual \underline{LE}arning} (shown in Fig.~\ref{fig1}) that pre-trains deep models using X-ray images collected from multiple body parts
subject to various tasks, and made contributions as follow.
%
%
%
%
\begin{enumerate}
    
    \item We study the problem of multi-dataset/multi-task SSL for X-ray images. To best of our knowledge, only few works have been done in related area~\cite{ref_article_1,ref_article_13}, especially by addressing \emph{data heterogeneity} (e.g., image sizes, resolutions, and gray-scale distributions), \emph{over-fitting} (e.g., to any one of the tasks), and \emph{catastrophic forgetting} (e.g., ejection of knowledge learned previously) in multi-dataset/multi-task learning settings.
    
    \item We present \TheName{} that pre-trains the backbone in multi-dataset/multi-task learning settings with task-specific heads, including Fully-Connected (FC) Layer, DeepLab-V3~\cite{ref_article_13-1}, and FasterRCNN~\cite{ref_article_13-2} for classification, segmentation, and abnormal detection tasks respectively. As shown in Fig.~\ref{fig1}, \TheName{} (1) adopts  MoCo to learn representations with a backbone network from multiple datasets, with pre-processing to tackle \emph{data heterogeneity}, and further (2) pre-trains the backbone to learn discriminative features with continual learning (CL) subject to multiple tasks, while avoiding over-fitting and ``catastrophic forgetting''~\cite{ref_article_3,ref_article_i4}. \TheName{} finally (3) fine-tunes the pre-trained backbone to adapt every task independently and separately.
    
    \item In this work, we collect 9 X-ray datasets to evaluate \TheName{}, where we pre-train and fine-tune the network to adapt all tasks using training subsets, and validate the performance using testing subsets. The experimental results show \TheName{} outperforms ImageNet/MoCo pre-trained backbones~\cite{ref_article_1} using both ResNet-18 and ResNet-50 as backbones. The comparisons confirm the \emph{proof-of-concept} of \TheName{}--self-supervised multi-task/dataset continual pre-training can boost performance of X-ray image analysis.
\end{enumerate}


\section{Methodologies}
In this section, we present the framework and algorithms design for \TheName.
As was shown in Fig.~\ref{fig1}, \TheName{} consists of three steps as follows.
\begin{enumerate}
    \item \underline{\emph{Multi-dataset momentum contrastive learning.}~(MD-MoCo)} Give multiple d-atasets of X-ray images collected from different body parts and based on different image resolution and gray-scale distributions, \TheName{} aggregates these datasets with pre-processing (e.g., resize, re-scale, and normalization), then adopts a MoCo-based SSL algorithm~\cite{ref_article_1} to pre-train the backbone networks and learn the representation of X-ray images within the aggregated dataset.
    
    \item \underline{\emph{Multi-task continual learning.}} Given the MD-MoCo pre-trained backbone and the datasets for different X-ray analysis tasks, e.g., pneumonia classification,  skeletal abnormality classification, and lung segmentation, \TheName{} leverages continual learning algorithms and further pre-trains a unified backbone network with alternating heads to learn the various sets of discriminative features subject to the different tasks.
    
    \item \underline{\emph{Fine-tuning with downstream tasks.}} Given the pre-trained backbone and every downstream X-ray image analysis task, \TheName{} fine-tunes and outputs a neural network using pre-trained weights as the initialization and its own task-specific head to fit the task independently. 
\end{enumerate}

%



\input{dataset-v1}
\subsection{Multi-Dataset Momentum Contrastive Learning (MD-MoCo)}
To pre-train the backbone with multiple datasets, \TheName{} collects and aggregates nine X-ray image datasets listed in Tab~\ref{tab1}. While these datasets have offered nearly 179,000 X-ray images covering cover several parts of the human body, including the chest, hand, elbow, finger, forearm, humerus, shoulder and wrist, \TheName{} makes non-trivial extension to adopt MoCo~\cite{ref_article_i3,ref_article_1} for \emph{multi-dataset momentum contrastive learning} as follows.

\paragraph{Re-sizing, Re-scaling, Normalization, and Aggregation}~The major challenge to work with multiple X-ray image datasets collected from different body part is the heterogeneity of images, including X-ray image resolutions and the distribution of gray-scales. To aggregate these datasets, several image pre-processing schemes have been used, where \TheName{} transforms and normalizes the gray-scale distribution of these datasets using the Z-score method with the mean of 122.786 and a standard deviation of 18.390. Further to fully utilize GPU for the resource-consuming MoCo algorithms, \TheName{} re-sizes all images into a 800$\times$500 resolution which balances the effectiveness and efficiency of deep learning.

\paragraph{MoCo-based Pre-training with Aggregated Datasets}~To utilize MoCo algorithm~\cite{ref_article_i3} for X-ray images pre-training, \TheName{} further extends MoCo-CXR~\cite{ref_article_1} with advanced settings on data augmentation strategies and initialization tricks. Specifically, while MoCo uses random data augmentation to generate contrastive views, \TheName{} disables random cropping, gaussian blurring, color/gray-scale jittering to preserve the semantic information for medical images. Furthermore, \TheName{} initializes the MoCo-based pre-training procedure with Kaiming's initialization~\cite{ref_article_23} to setup the convolution layers, so as to ensure the stability of back-propagation in contrastive learning.

Note that, in this work, the training sets of all nine datasets have been used for \emph{multi-dataset momentum contrastive learning}, while four of them with specific X-ray image analysis tasks are further used for \emph{multi-task continual learning}.

\subsection{Multi-Task Continual Learning}~
To further pre-train the backbone subject to task-specific representations, \TheName{} adopts the continual learning (CL)~\cite{ref_article_25} with four X-ray image analysis tasks, including pneumonia classification from Pneumonia~\cite{ref_article_18}, skeletal abnormality classification from MURA~\cite{ref_article_21}, lung segmentation from Chest Xray Masks and Labels~\cite{ref_article_16}, and Tuberculosis(TB) detection from TBX~\cite{ref_article_22}. Specifically, \TheName{} extends the vanilla CL for neural network, which iterates the training procedures of the backbone network with alternating task-specific heads subject to tasks, with two major advancements as follows.

\paragraph{Cyclic and Reshuffled Learning Schedule}~\TheName{} splits the continual learning procedure into 10 \textbf{\em rounds} of learning process, where each round of learning process \textbf{\em iterates} the 4 learning tasks one-by-one and each iterate of learning task trains the backbone network with 1 epoch using a task-specific head, i.e., Fully-Connected (FC) Layer for classification tasks, DeepLab-V3~\cite{ref_article_13-1} for segmentation tasks, and FasterRCNN~\cite{ref_article_13-2} for detection tasks. Furthermore, to avoid overfitting to any task, \TheName{} reshuffle the order of tasks in every round of learning process, and adopts Cosine annealing learning rate schedule as follow.
    \begin{equation}
    \eta_{t}=\eta_{\min }^{i}+\frac{1}{2}\left(\eta_{\max }-\eta_{\min }\right)\left(1+\cos \left(\frac{t}{T} \cdot 2\pi\right)\right)
    \end{equation}
    where $\eta_{t}$ refers to the learning rate of the $t^{th}$ iteration, $\eta_\mathrm{max}$ and $\eta_\mathrm{min}$ and $T_i$ refer to the maximal and minimal learning rates, and $T$ refers to the total number of iterations within a period of cyclic learning rate schedule.

\paragraph{Cross-Task Memorization with Explicit Bias}~Yet another challenge of multi-task CL is ``catastrophic forgetting", where the backbone would ``forget'' the knowledge learned from the previous iterates. To solve the problem, \TheName{} leverages a knowledge transfer regularization derived from L$^2$-SP~\cite{ref_article_3}. In each iterate of learning task, given the pre-trained model obtained from previous iterates, \TheName{} sets the pre-trained weights as $\boldsymbol{w}^0_{\mathcal{S}}$ and trains the backbone using the following loss,
    \input{l2-sp}
where $\boldsymbol{w}_{\mathcal{S}}$ is the learning outcome and $\alpha$ is the hyper-parameter. Thus, above regularization constrains the distance between the learning outcome and the backbone trained by the previous iterates.

\subsection{Fine-tuning on Downstream Tasks}
Finally, given the backbone pre-trained using above two steps, \TheName{} fine-tunes the backbone on each of the four tasks independent and separately. Again, \TheName{} connects the pre-trained backbone with Fully-Connected (FC) Layer for classification tasks, DeepLab-V3~\cite{ref_article_13-1} for segmentation tasks, and FasterRCNN~\cite{ref_article_13-2} for detection tasks. Finally, \TheName{} employs the standard settings, such as vanilla weight decay as stability regularization and step-based decay of learning rate schedule, for such fine-tuning.


\input{pne-mura-res-tab-v2}

\section{Experiment}
In this section, we present our experimental results to confirm the effectiveness of \TheName{} on various tasks. 

\paragraph{Experiment Setups} In this experiments, we include performance comparisons on pneumonia classification (Pneumonia)~\cite{ref_article_18}, skeletal abnormality classification (MURA)~\cite{ref_article_21}, lung segmentation (Lung)~\cite{ref_article_16}, and  tuberculosis detection (TBX)~\cite{ref_article_22} tasks. We use the training sets of these datasets/tasks to pre-train and fine-tune the network with \TheName{} or other baseline algorithms, tune the hyper-parameters using validation sets, and report results on testing datasets.

\begin{figure}[t]
\centering
\includegraphics[width=\textwidth]{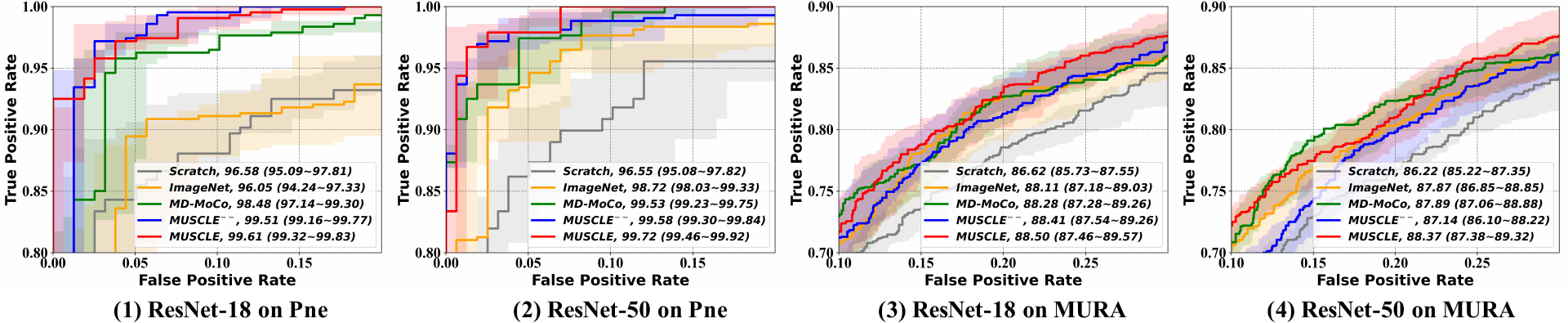}
\caption{Receiver Operating Characteristic (ROC) Curves with AUC value (95\%CI) on pneumonia classification~(Pne) and skeletal abnormality classification~(MURA)} 
\label{fig2}
\end{figure}

As the goal of \TheName{} is to pre-train backbones under multi-task/dataset settings, we propose several baselines for comparisons as follows. \textbf{Scratch}: the models are all initialized using Kaiming's random initialization~\cite{ref_article_23} and fine-tuned on the target datasets (introduced in Section 2.3). \textbf{ImageNet}: the models are initialized using the officially-released weights pre-trained by the ImageNet dataset and fine-tuned on the target datasets. \textbf{MD-MoCo}: the models are pre-trained using \emph{multi-dataset MoCo} (introduced in Section 2.2) and fine-tuned accordingly; we believe {MD-MoCo} is one of our proposed methods and can prove the concepts, as {MD-MoCo} extends both vanilla MoCo~\cite{ref_article_2} and MoCo-CXR~\cite{ref_article_1} with additional data pre-processing methods to tackle the multi-dataset issues. \TheName{}$^{--}$: all models are pre-trained and fine-tuned with \TheName{} but with \emph{Cross-Task Memorization} and \emph{Cyclic and Reshuffled Learning Schedule} turned off. All models here are built with ResNet-18 and ResNet-50.

To compare these algorithms, we evaluate and compare \emph{Accuracy} (Acc.), \emph{Area under the Curve} (AUC), \emph{Sensitivity} (Sen.), \emph{Specificity} (Sep.) and \emph{Receiver Operating Characteristic (ROC) Curve} for Pneumonia and MURA classification tasks, \emph{Dice Similarity Coefficient} (Dice), \emph{mean Intersection over Union} (mIoU) for lung segmentation (Lung) task, and \emph{mean Average Precision} (mAP), \emph{Average Precision of Active TB} ($\mathrm{AP}_\mathrm{active}$) and \emph{Average Precision of 
Latent TB} ($\mathrm{AP}_\mathrm{latent}$) for tuberculosis detection (TBX), all of them at the IoU threshold of 0.5. AUC values are reported with 95\% confidence intervals estimated through bootstrapping with 100 independent trials on testing sets.

\begin{figure}[t]
\centering
\includegraphics[width=\textwidth]{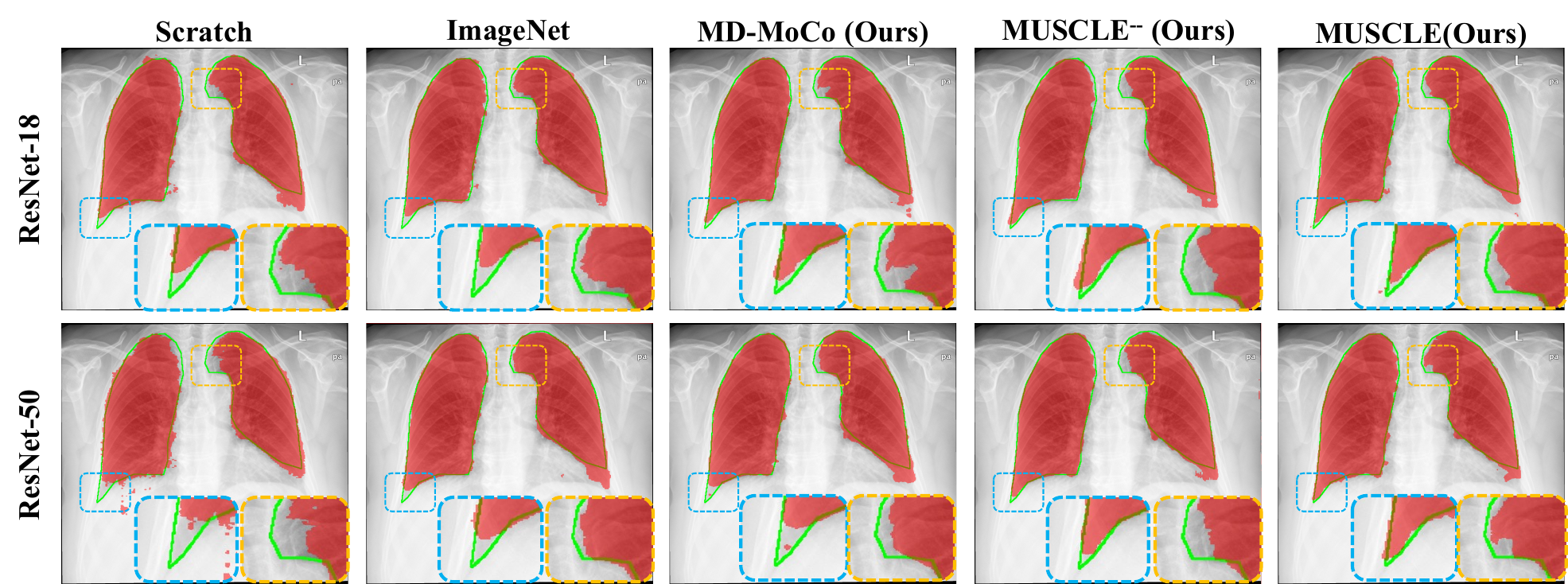}
\caption{Results using various pre-training algorithms on lung segmentation, where green lines indicate ground truth and red areas indicate model prediction results, and the blue and orange boxes cover regions of the main differences in these results} \label{fig3}
\end{figure}

\paragraph{Overall Comparisons} We present the results on Pneumonia and MURA classification tasks in Table~\ref{tab2} and Fig.~\ref{fig2}, where we compare the performance of \TheName{} with all baseline algorithms and plot the ROC curve. Results show that our proposed methods, including \TheName{}, \TheName{}$^{--}$ and MD-MoCo can significantly outperform the one based on ImageNet pre-trained models and Scratch. Furthermore, in terms of overall accuracy (Acc. and AUC), \TheName{} outperforms MD-MoCo and \TheName{}$^{--}$ in all cases, due to its advanced continual learning settings. However, we can still observe a slightly lower sensitivity on the Pneumonia classification task and a slightly lower specificity on the MURA delivered by \TheName{}. Similar observations could be found in Table~\ref{tab3} and Fig.~\ref{fig3}, where we report the performance on lung segmentation and TB detection tasks with examples of lung segmentation plotted. 

\paragraph{More Comparisons} Please note that MD-MoCo indeed surpasses \textbf{MoCo}~\cite{ref_article_2} and \textbf{MoCo-CXR}~\cite{ref_article_1} (state of the art in X-rays pre-training), in terms of performance, as it uses multiple datasets for pre-training and solves the data heterogeneity problem (the performance of MoCo might be even worse than ImageNet-based pre-training, if we don't unify the gray-scale distributions of X-ray images collected from different datasets). We also have tried to replace MoCo with \textbf{SimCLR}~\cite{ref_article_5}, the performance of SimCLR for X-ray pre-training is even worse, e.g., 87.52\% (\textcolor{red}{9.57\%$\downarrow$}) Acc. for pneumonia classification.

Furthermore, though \TheName{} is proposed as a \emph{``proof-of-concept''} solution and has not been optimized for any single medical imaging task, e.g., using U-Net for segmentation~\cite{ref_article_26}, the overall performance of \TheName{} is still better than many recent works based on the same datasets. For example, \emph{Stephen et al.~(2019)}~\cite{ref_article_27} reports a 93.73\% (\textcolor{red}{4.42\%$\downarrow$}) accuracy for the same pneumonia classification task and \emph{Li et al.~(2021)}~\cite{ref_article_28} reports a  94.64\% Dice (\textcolor{red}{0.73\%$\downarrow$}) for the lung segmentation. For MURA, \emph{Bozorgtabar et al.~(2020)}~\cite{ref_article_29} reports an AUC of 82.45\%  (\textcolor{red}{6.05\%$\downarrow$}), while \emph{Liu et al.~(2020)}~\cite{ref_article_22} reports a 58.70\% (\textcolor{red}{4.76\%$\downarrow$}) $\mathrm{AP}_\mathrm{active}$ and a 9.60\% (\textcolor{red}{2.61\%$\downarrow$}) $\mathrm{AP}_\mathrm{latent}$ for TBX based on FasterRNN. The advantages of \TheName{} demonstrate the feasibility of using multiple tasks/datasets to pre-train the backbone with SSL and CL. For more results, please refer to the appendix.

\paragraph{Ablation Studies} The comparisons between Scratch versus MD-MoCo, between MD-MoCo versus \TheName{}$^{--}$, and between \TheName{}$^{--}$ versus \TheName{} confirm the contributions made by each step of our algorithms. Frankly speaking, in many cases, MD-MoCo achieved the most performance improvement (compared to ImageNet or Scratch), while continual learning without \emph{Cyclic \& Reshuffled Learning Schedule} and \emph{Cross-Task Memorization} may even make \TheName{}$^{--}$ performs worse, due to over-fitting or catastrophic forgetting. However, \TheName{}, 
pipelining MD-MoCo and advanced continual learning strategies, successfully defends her dominant position and outperforms other algorithms in most cases.

\begin{table}[t]
\centering
\caption{Performance Comparisons for Lung Segmentation(Lung) and TB Dtection(TBX) using various Pre-training Algorithms.}\label{tab3}
\begin{tabular}{p{50 pt}p{50 pt}p{45 pt}p{45 pt}p{45 pt}p{45 pt}p{45 pt}}
\hline
\multirow{2}*{\bfseries Backbones} & \multirow{2}*{\bfseries Pre-train} & \multicolumn{2}{c}{\bfseries Lung} & \multicolumn{3}{c}{\bfseries TBX}\\
\cmidrule(r){3-4}  \cmidrule(r){5-7}
 & & \multicolumn{1}{c}{\bfseries Dice} & \multicolumn{1}{c}{\bfseries mIoU} & \multicolumn{1}{c}{\bfseries mAP} & \multicolumn{1}{c}{\bfseries AP{\fontsize{3}{4}\selectfont Active}} & \multicolumn{1}{c}{\bfseries AP{\fontsize{3}{4}\selectfont Latent}}\\
\hline
\multirow{5}*{ResNet-18} 
& Scratch & \multicolumn{1}{r}{95.24} & \multicolumn{1}{r}{94.00}& \multicolumn{1}{r}{30.71} & \multicolumn{1}{r}{56.71} & \multicolumn{1}{r}{4.72}\\
& ImageNet & \multicolumn{1}{r}{95.26} & \multicolumn{1}{r}{94.10} & \multicolumn{1}{r}{29.46} & \multicolumn{1}{r}{56.27} & \multicolumn{1}{r}{2.66}\\
& MD-MoCo & \multicolumn{1}{r}{95.31} & \multicolumn{1}{r}{94.14} & \multicolumn{1}{r}{36.00} & \multicolumn{1}{r}{\bfseries 67.17} & \multicolumn{1}{r}{4.84}\\
& \TheName{}$^{--}$ & \multicolumn{1}{r}{95.14} & \multicolumn{1}{r}{93.90} & \multicolumn{1}{r}{34.70} & \multicolumn{1}{r}{63.43} & \multicolumn{1}{r}{5.97}\\
& \TheName{} & \multicolumn{1}{r}{\bfseries 95.37} & \multicolumn{1}{r}{\bfseries 94.22} & \multicolumn{1}{r}{\bfseries 36.71} & \multicolumn{1}{r}{64.84} & \multicolumn{1}{r}{\bfseries 8.59}\\
\hline
\multirow{5}{*}{ResNet-50}
& Scratch & \multicolumn{1}{r}{93.52} & \multicolumn{1}{r}{92.03} & \multicolumn{1}{r}{23.93} & \multicolumn{1}{r}{44.85} & \multicolumn{1}{r}{3.01}\\
& ImageNet & \multicolumn{1}{r}{93.77} & \multicolumn{1}{r}{92.43} & \multicolumn{1}{r}{35.61} & \multicolumn{1}{r}{58.81} & \multicolumn{1}{r}{12.42}\\
& MD-MoCo & \multicolumn{1}{r}{94.33} & \multicolumn{1}{r}{93.04} & \multicolumn{1}{r}{36.78} & \multicolumn{1}{r}{\bfseries 64.37} & \multicolumn{1}{r}{9.18}\\
& \TheName{}$^{--}$ & \multicolumn{1}{r}{95.04} & \multicolumn{1}{r}{93.82} & \multicolumn{1}{r}{35.14} & \multicolumn{1}{r}{57.32} & \multicolumn{1}{r}{\bfseries 12.97}\\
& \TheName{} & \multicolumn{1}{r}{\bfseries 95.27} & \multicolumn{1}{r}{\bfseries 94.10} & \multicolumn{1}{r}{\bfseries 37.83} & \multicolumn{1}{r}{63.46} & \multicolumn{1}{r}{12.21}\\
\hline
\end{tabular}
\end{table}

\section{Discussion and Conclusion}
In this work, we present \TheName{} a self-supervised continual learning pipeline that pre-trains deep neural networks using multiple X-ray image datasets collected from different body parts, e.g., hands, chests, bones and etc., for multiple X-ray analysis tasks, e.g., TB detection, lung segmentation, and skeletal abnormality classification. \TheName{} proposes \emph{multi-dataset momentum contrastive learning} (MD-MoCo) and \emph{multi-task continual learning} to tackle the data heterogeneity, over-fitting, and catastrophic forgetting problems in pre-training, and finally fine-tunes the network to adapt every task independently and separately. Experiment results on 9 X-ray image datasets show \TheName{} outperforms other pre-training methods, including ImageNet-based and MoCo-based~\cite{ref_article_1} solutions. We do acknowledge that \TheName{} might NOT be an optimized solution for any specific task in this study, we however claim \TheName{} as a \emph{``proof-of-concept''} that demonstrates the feasibility of using multiple datasets/tasks to pre-train X-ray models with advanced strategies of self-supervised continual learning.



%
%
%
%

\end{document}

%% file: dataset-v1.tex
\newcommand{\tabincell}[2]{\begin{tabular}{@{}#1@{}}#2\end{tabular}}
\begin{table}\centering
\caption{An Overview of Nine Publicly Available X-ray Image Datasets}
\begin{tabular}{p{110 pt}p{50 pt}p{50 pt}p{50 pt}p{50 pt}p{50 pt}p{50 pt}}
\hline
\multicolumn{1}{c}{\bfseries Datasets} & \multicolumn{1}{c}{\bfseries Body Part} & \multicolumn{1}{c}{\bfseries Task} & \multicolumn{1}{c}{\bfseries Train} & \multicolumn{1}{c}{\bfseries Valid} & \multicolumn{1}{c}{\bfseries Test} & \multicolumn{1}{c}{\bfseries Total}\\
\hline
\multicolumn{7}{c}{Only Used for the first step (MD-MoCo) of \TheName{}}\\\hline
NIHCC~\cite{ref_article_15} & \multicolumn{1}{c}{Chest} & \multicolumn{1}{c}{N/A} & \multicolumn{1}{r}{112,120} &  \multicolumn{1}{r}{N/A} & \multicolumn{1}{r}{N/A} & \multicolumn{1}{r}{112,120}\\
China-Set-CXR~\cite{ref_article_16} & \multicolumn{1}{c}{Chest} & \multicolumn{1}{c}{N/A} & \multicolumn{1}{r}{661} & \multicolumn{1}{r}{N/A} & \multicolumn{1}{r}{N/A} & \multicolumn{1}{r}{661}\\
Montgomery-Set-CXR~\cite{ref_article_16} & \multicolumn{1}{c}{Chest} & \multicolumn{1}{c}{N/A} & \multicolumn{1}{r}{138} & \multicolumn{1}{r}{N/A} & \multicolumn{1}{r}{N/A} & \multicolumn{1}{r}{138}\\
Indiana-CXR~\cite{ref_article_17} & \multicolumn{1}{c}{Chest} & \multicolumn{1}{c}{N/A} & \multicolumn{1}{r}{7,470} & \multicolumn{1}{r}{N/A} & \multicolumn{1}{r}{N/A} & \multicolumn{1}{r}{7,470}\\
RSNA Bone Age~\cite{ref_article_20} & \multicolumn{1}{c}{Hand} & \multicolumn{1}{c}{N/A} & \multicolumn{1}{r}{10,811} & \multicolumn{1}{r}{N/A} & \multicolumn{1}{r}{N/A} & \multicolumn{1}{r}{10,811}\\\hline
\multicolumn{7}{c}{Used for all three steps of \TheName{}}\\\hline
Pneumonia~\cite{ref_article_18} & \multicolumn{1}{c}{Chest} & \multicolumn{1}{c}{\tabincell{c}{Classification}} & \multicolumn{1}{r}{4,686} & \multicolumn{1}{r}{585} & \multicolumn{1}{r}{585} & \multicolumn{1}{r}{5,856}\\
MURA~\cite{ref_article_21} & \multicolumn{1}{c}{\tabincell{c}{Various Bones}} & \multicolumn{1}{c}{\tabincell{c}{Classification}} & \multicolumn{1}{r}{32,013} & \multicolumn{1}{r}{3,997} & \multicolumn{1}{r}{3,995} & \multicolumn{1}{r}{40,005}\\
Chest Xray Masks..~\cite{ref_article_16} & \multicolumn{1}{c}{Chest} & \multicolumn{1}{c}{\tabincell{c}{Segmentation}} & \multicolumn{1}{r}{718} & \multicolumn{1}{r}{89} & \multicolumn{1}{r}{89} & \multicolumn{1}{r}{896}\\
TBX~\cite{ref_article_22} & \multicolumn{1}{c}{Chest} & \multicolumn{1}{c}{\tabincell{c}{Detection}} & \multicolumn{1}{r}{640} & \multicolumn{1}{r}{80} & \multicolumn{1}{r}{80} & \multicolumn{1}{r}{800}\\
\hline
{\bfseries Total} & \multicolumn{1}{c}{N/A} & \multicolumn{1}{c}{N/A} & \multicolumn{1}{r}{169,257} & \multicolumn{1}{r}{4,751} & \multicolumn{1}{r}{4,749} & \multicolumn{1}{r}{178,757}\\
\hline
\end{tabular}
\label{tab1}
\end{table}

%% file: l2-sp.tex
\begin{figure}[ht]
\begin{equation}
\label{eq:ab_flushing2}
\Omega(\boldsymbol{w}_{\mathcal{S}})=
\color{purple}
\overbrace{ \color{black}
{\alpha} \left\|
\tikzmarknode{ws}{\highlight{xkcdHunterGreen}{\color{black}$\boldsymbol{w}_{\mathcal{S}}$}}
-
\tikzmarknode{ws0}{\highlight{Bittersweet}{\color{black}$\boldsymbol{w}_{\mathcal{S}}^{0}$}}
\right\|_{2}^{2}
}^{\substack{\text{\sf \scriptsize \textcolor{purple!85}{ distance to the model
	}} \\ \text{\sf \scriptsize \textcolor{purple!85}{ pre-trained by previous iterates.
}} } }
\color{black} + \quad
\color{NavyBlue}
\overbrace{ \color{black}
{(1-\alpha)}\left\|
\tikzmarknode{ws-}{\highlight{xkcdHunterGreen}{\color{black}$\boldsymbol{w}_{\mathcal{S}}$}}
\right\|_{2}^{2}
}^{\substack{\text{\sf \scriptsize \textcolor{NavyBlue!85}{Ridge-based
	}} \\ \text{\sf \scriptsize \textcolor{NavyBlue!85}{ regularization.
}} } }
\color{black}
\end{equation}
\begin{tikzpicture}[overlay,remember picture,>=stealth,nodes={align=left,inner ysep=1pt},<-]
\path (ws.north) ++ (-2.0,-1.8em) node[anchor=north west,color=xkcdHunterGreen!85] (wstext){\textsf{\scriptsize outcome of CL}};
\draw [color=xkcdHunterGreen](ws.south) |- ([xshift=-0.3ex,color=xkcdHunterGreen]wstext.south west);
\path (ws0.north) ++ (0.1,-2.2em) node[anchor=north west,color=Bittersweet!85] (ws0text){\textsf{\scriptsize pre-trained by previous iterates}};
\draw [color=Bittersweet](ws0.south) |- ([xshift=-0.3ex,color=Bittersweet]ws0text.south east);
\end{tikzpicture}
\end{figure}

%% file: pne-mura-res-tab-v2.tex
\begin{table}[t]
\centering
\caption{Performance Comparisons for Pneumonia Classification (Pneumonia) and Skeletal Abnormality Classification (MURA) using various Pre-training Algorithms.}\label{tab2}
\begin{tabular}{p{60pt}p{50pt}p{50pt}p{40pt}p{40pt}p{40pt}p{70pt}}
\hline
\multicolumn{1}{c}{\bf Datasets} & \multicolumn{1}{c}{\bf Backbones} & \multicolumn{1}{c}{\bf Pre-train} & \multicolumn{1}{c}{\bfseries Acc.} & \multicolumn{1}{c}{\bfseries Sen.} & \multicolumn{1}{c}{\bfseries Spe.} & \multicolumn{1}{c}{\bfseries AUC 
(95\%CI) 
}\\
\hline
\multirow{10}*{\bf Pneumonia}
& \multirow{5}*{ResNet-18}
& Scratch & \multicolumn{1}{r}{91.11} & \multicolumn{1}{r}{93.91} & \multicolumn{1}{r}{83.54} & \multicolumn{1}{r}{96.58 (95.09-97.81)}\\
& & ImageNet & \multicolumn{1}{r}{90.09} & \multicolumn{1}{r}{93.68} & \multicolumn{1}{r}{80.38} & \multicolumn{1}{r}{96.05 (94.24-97.33)}\\
& & MD-MoCo & \multicolumn{1}{r}{96.58} & \multicolumn{1}{r}{97.19} & \multicolumn{1}{r}{94.94} & \multicolumn{1}{r}{98.48 (97.14-99.30)}\\
& & \TheName{}$^{--}$ & \multicolumn{1}{r}{96.75} &  \multicolumn{1}{r}{\bfseries 97.66} &  \multicolumn{1}{r}{94.30} & \multicolumn{1}{r}{99.51 (99.16-99.77)}\\
& & \TheName{} & \multicolumn{1}{r}{\bfseries 97.26} & \multicolumn{1}{r}{97.42} & \multicolumn{1}{r}{\bfseries 96.84} & \multicolumn{1}{r}{\bfseries 99.61 (99.32-99.83)}\\
\cmidrule(r){2-7}
& \multirow{5}{*}{ResNet-50}
& Scratch & \multicolumn{1}{r}{91.45} & \multicolumn{1}{r}{92.51} & \multicolumn{1}{r}{88.61} & \multicolumn{1}{r}{96.55 (95.08-97.82)}\\
& & ImageNet & \multicolumn{1}{r}{95.38} & \multicolumn{1}{r}{95.78} & \multicolumn{1}{r}{94.30} & \multicolumn{1}{r}{98.72 (98.03-99.33)}\\
& & MD-MoCo & \multicolumn{1}{r}{97.09} & \multicolumn{1}{r}{\bfseries 98.83} & \multicolumn{1}{r}{92.41} & \multicolumn{1}{r}{99.53 (99.23-99.75)}\\
& & \TheName{}$^{--}$ & \multicolumn{1}{r}{96.75} & \multicolumn{1}{r}{98.36} & \multicolumn{1}{r}{92.41} & \multicolumn{1}{r}{99.58 (99.30-99.84)}\\
& & \TheName{} & \multicolumn{1}{r}{\bfseries 98.12} & \multicolumn{1}{r}{98.36} & \multicolumn{1}{r}{\bfseries 97.47} & \multicolumn{1}{r}{\bfseries 99.72 (99.46-99.92)}\\
\hline
\multirow{10}*{\bf MURA}
& \multirow{5}*{ResNet-18}
& Scratch & \multicolumn{1}{r}{81.00} & \multicolumn{1}{r}{68.17} & \multicolumn{1}{r}{\bfseries 89.91} & \multicolumn{1}{r}{86.62 (85.73-87.55)}\\
& & ImageNet & \multicolumn{1}{r}{81.88} & \multicolumn{1}{r}{73.49} & \multicolumn{1}{r}{87.70} & \multicolumn{1}{r}{88.11 (87.18-89.03)}\\
& & MD-MoCo & \multicolumn{1}{r}{82.48} & \multicolumn{1}{r}{72.27} & \multicolumn{1}{r}{89.57} & \multicolumn{1}{r}{88.28 (87.28-89.26)}\\
& & \TheName{}$^{--}$ & \multicolumn{1}{r}{82.45} &  \multicolumn{1}{r}{74.16} &  \multicolumn{1}{r}{88.21} & \multicolumn{1}{r}{88.41 (87.54-89.26)}\\
& & \TheName{} & \multicolumn{1}{r}{\bfseries 82.62} & \multicolumn{1}{r}{\bfseries 74.28} & \multicolumn{1}{r}{88.42} & \multicolumn{1}{r}{\bfseries 88.50 (87.46-89.57)}\\
\cmidrule(r){2-7}
& \multirow{5}{*}{ResNet-50}
& Scratch & \multicolumn{1}{r}{80.50} & \multicolumn{1}{r}{65.42} & \multicolumn{1}{r}{90.97} & \multicolumn{1}{r}{86.22 (85.22-87.35)}\\
& & ImageNet & \multicolumn{1}{r}{81.73} & \multicolumn{1}{r}{68.36} & \multicolumn{1}{r}{\bfseries 91.01} & \multicolumn{1}{r}{87.87 (86.85-88.85)}\\
& & MD-MoCo & \multicolumn{1}{r}{82.35} & \multicolumn{1}{r}{73.12} & \multicolumn{1}{r}{88.76} & \multicolumn{1}{r}{87.89 (87.06-88.88)}\\
& & \TheName{}$^{--}$ & \multicolumn{1}{r}{81.10} & \multicolumn{1}{r}{69.03} & \multicolumn{1}{r}{89.48} & \multicolumn{1}{r}{87.14 (86.10-88.22)}\\
& & \TheName{} & \multicolumn{1}{r}{\bfseries 82.60} & \multicolumn{1}{r}{\bfseries 74.53} & \multicolumn{1}{r}{88.21} & \multicolumn{1}{r}{\bfseries 88.37 (87.38-89.32)}\\
\hline
\end{tabular}
\end{table}